\documentclass{article}

\PassOptionsToPackage{numbers}{natbib}
\usepackage[preprint]{neurips_2022}

\usepackage[utf8]{inputenc} % allow utf-8 input
\usepackage[T1]{fontenc}    % use 8-bit T1 fonts
\usepackage{hyperref}       % hyperlinks
\usepackage{url}            % simple URL typesetting
\usepackage{booktabs}       % professional-quality tables
\usepackage{amsfonts}       % blackboard math symbols
\usepackage{nicefrac}       % compact symbols for 1/2, etc.
\usepackage{microtype}      % microtypography
\usepackage{xcolor}         % colors
\usepackage{titlesec} % modify spacing of sections

\titlespacing*{\section}{1ex}{0.2ex}{0.1ex}
\titlespacing*{\subsection}{1ex}{0.2ex}{0.1ex}

\usepackage{graphicx}
\usepackage[export]{adjustbox}
\usepackage[aboveskip=1pt,labelfont=bf,labelsep=period,justification=raggedright,singlelinecheck=off]{caption}

\usepackage{amsmath,amssymb}
\DeclareMathOperator*{\argmax}{argmax} % thin space, limits underneath in displays

\title{Efficient And Robust Multi-Task Learning In The Brain With Modular Latent Primitives}
\author{
  Christian D. Márton\\
  Dept of Neuroscience \& Friedman Brain Institute\\
  Icahn School of Medicine at Mount Sinai\\
  New York, NY 10029 \\
  \texttt{cdmarton@gmail.com} \\
  \And
  Leo Gagnon\\
  Dep. of Computer Science\\
  Université de Montréal\\
  MILA - Québec AI Institute\\
  \texttt{leo.gagnon@umontreal.ca}\\
  \And
  Kanaka Rajan\footnote[1]{}\\
  Dept of Neuroscience \& Friedman Brain Institute\\
  Icahn School of Medicine at Mount Sinai\\
  New York, NY 10029\\
  \texttt{kanaka.rajan@mssm.edu}\\
  *Shared senior authorship
  \And
  Guillaume Lajoie\footnote[1]{}\\
  Dep. of Mathematics and Statistics\\
  Université de Montréal\\
  MILA - Québec AI Institute\\
  \texttt{g.lajoie@umontreal.ca}\\
  *Shared senior authorship
  }

\begin{document}

\maketitle

\begin{abstract}
Biological agents do not have infinite resources to learn new things. For this reason, a central aspect of human learning is the ability to recycle previously acquired knowledge in a way that allows for faster, less resource-intensive acquisition of new skills. In spite of that, how neural networks in the brain leverage existing knowledge to learn new computations is not well understood. In this work, we study this question in artificial recurrent neural networks (RNNs) trained on a corpus of commonly used neuroscience tasks. Combining brain-inspired inductive biases we call functional and structural, we propose a system that learns new tasks by building on top of pre-trained latent dynamics organised into separate  recurrent modules. These modules, acting as prior knowledge acquired previously through evolution or development, are pre-trained on the statistics of the full corpus of tasks so as to be independent and maximally informative. The resulting model, we call a Modular Latent Primitives (MoLaP) network, allows for learning multiple tasks while keeping parameter counts, and updates, low. We also show that the skills acquired with our approach are more robust to a broad range of perturbations compared to those acquired with other multi-task learning strategies, and that generalisation to new tasks is facilitated. This work offers a new perspective on achieving efficient multi-task learning in the brain, illustrating the benefits of leveraging pre-trained latent dynamical primitives.
% Biological agents do not have infinite resources to learn new things. For this reason, a central aspect of human learning is the ability to recycle previously acquired knowledge in a way that allows for faster, less resource-intensive acquisition of new skills. In spite of that, how neural networks in the brain leverage existing knowledge to learn new computations is not well understood. In this work, we study this question in artificial recurrent neural networks (RNNs) trained on a corpus of commonly used neuroscience tasks. Combining brain-inspired inductive biases we call \textit{functional} and \textit{structural}, we propose a system that learns new tasks by building on top of pre-trained latent dynamics organised into separate recurrent modules. These modules, acting as prior knowledge acquired previously through evolution or development, are pre-trained on the statistics of the full corpus of tasks so as to be independent and maximally informative. The resulting model, we call a Modular Latent Primitives (MoLaP) network, allows for learning multiple tasks while keeping parameter counts, and updates, low. We also show that the skills acquired with our approach are more robust to a broad range of perturbations compared to those acquired with other multi-task learning strategies, and that generalisation to new tasks is facilitated. This work offers a new perspective on achieving efficient multi-task learning in the brain by illustrating the benefits of leveraging pre-trained primitive modules.
\end{abstract}

\section{Introduction}\label{introduction}
% RNNs can be used to solve cognitive tasks, but from scratch and on only one task unlike humans
A large body of work has shown that computations in the brain --- implemented by the collective dynamics of large populations of neurons --- can be modeled with recurrent artificial neural networks (RNNs) \cite{marton_learning_2020,richards_deep_2019,yang_task_2018,remington_dynamical_2018,kell_task-optimized_2018,zeng_continuous_2018,chaisangmongkon_computing_2017,rajan_recurrent_2016,sussillo_neural_2015,mante_context-dependent_2013,sussillo_opening_2013,barak_fixed_2013,buonomano_state-dependent_2009,sussillo_generating_2009}. Along these lines, many investigations of the mechanisms underlying computations in the brain have been performed by training RNNs on tasks inspired by neuroscience experiments. In particular, recent studies \cite{duncker_organizing_2020,yang_task_2018} have proposed ways in which RNNs could learn multiple tasks while reusing parameters like biological agents do. However, in all of them, models start with no pre-existing knowledge about the world and need to learn all the shared parameters incrementally by seeing each task, one by one. Notably, this scenario, often studied in deep learning under \textit{continual learning}, is not exactly one in which biological agents usually operate.

%This has lead to insights into the computational mechanisms underlying the execution of multiple different tasks across various domains. In these studies, however, RNNs are typically trained from scratch on a unique task at the expense of long training duration. This is unlike biological agents, who are able to learn new tasks efficiently by making use of pre-existing knowledge previously acquired \cite{lake_2017}. 

Indeed, when faced with a new task, animals have access to varying amounts of pre-existing \textit{inductive biases} acquired over large timescales : evolutionary \cite{dominici_newborn_2011,richards_deep_2019} as well as those within their lifetime \cite{botvinick_reinforcement_2019, wang_metarl_2018}. These biases are assumptions about the solution of future tasks based on heredity or experience : something really useful for biological agents who often inhabit particular ecological niches \cite{carscadden_niche_2020} where a substantial structure is shared between tasks. In this work, we explore how the common statistics of a corpus of tasks --- standing in for the knowledge acquired through long timescales --- could be encoded  into an RNN in a biologically inspired way, and then leveraged to enable learning multiple tasks sequentially more efficiently than in traditional continual learning scenarios.
One way in which inductive biases can be expressed in the brain is in terms of neural architecture or \textit{structure} \cite{richards_deep_2019}. Neural networks in the brain may come with a pre-configured wiring diagram, encoded in the genome \cite{zador_critique_2019}. Such \textit{structural} pre-configurations can put useful constraints on the learning problem by forcing the representations to have a desirable structure.  For example, a neural network made up of different specialized modules encourages the learning of independent and reusable knowledge, which leads to better generalization \cite{pearl_causality_2009}. Moreover, modularity is known to increase the overall robustness of networks since it prevents local perturbation from spreading and affecting the dynamics too much \cite{chen_robustness_2021}. Previous neuroscience studies have shown that the brain exhibits specializations--and thus, modularity--across various scales \cite{dubreuil_structure_2021,sporns_modular_2016,wang_brain_2016,goulas_strength_2015,murray_hierarchy_2014,meunier_modular_2010}. For example, visual information is primarily represented in the visual cortices, auditory information in the auditory cortices,  decision-making in the prefrontal cortex--each of these regions being further comprised of cell clusters processing particular kinds of information \cite{marton_learning_2020,chaisangmongkon_computing_2017,mante_context-dependent_2013}. 

%Another type of inductive bias is functional, likely acquired over lifetime. An example of this are task sets, steretyped dynamics across neural ensembles in the motor domain, etc.
Another way in which inductive biases may be expressed in the brain, beyond structure, is in terms of \textit{function} : the specific neural dynamics instantiated in particular groups of neurons. Indeed, typical computations can be learned once and then recombined in subsequent tasks to enable more efficient learning; there is evidence that the brain implements such stereotyped dynamics \cite{domenech_executive_2015,sakai_task_2008}. Mixing in \textit{structural} inductive biases, these generic functional dynamics are often local to a certain group of neurons, forming modules \cite{perich_rethinking_2020,sporns_modular_2016,billeh_revealing_2014,power_functional_2011,meunier_modular_2010} that can perform different aspects of a task \cite{koay_sequential_2019, yang_task_2018, chaisangmongkon_computing_2017}. Thus, a network that implements modular functional \textit{primitives}, or \textit{basis}-computations, could offer dynamics useful for training multiple tasks. 

%In the motor domain, for instance, neural activity was found to exhibit shared structure across tasks  \cite{gallego_cortical_2018}. 
%and theoretical work suggests the number of different neural sub-populations influences the set of dynamics a neural network is able to perform \cite{beiran_shaping_2020}. 

% Our contribution. Structural + Functional biases in the form of pre-trained latent modules
Here, we investigate the advantages of combining \textit{structural} and \textit{functional} inductive biases with a modular RNN for learning multiple tasks efficiently. Specifically, our model learns a corpus of 9 different tasks from the neuroscience literature by building on top of shared, frozen latent modules. These modules are pre-trained to produce simple computations that constitute shared operations of the task ensemble. We achieve this by requiring modules to pre-learn leading canonical components of transformations required by the corpus as a whole, thus producing a provably disentangled and maximally informative latent representation (i.e., explaining the most covariance between input and outputs). Importantly, we conceptualize the pre-trained dynamics as being acquired through  evolution \cite{dominici_newborn_2011} or another biologically instantiated meta-learning process \cite{botvinick_reinforcement_2019,wang_metarl_2018}, providing a biologically plausible way to access the shared statistics of many tasks encountered throughout the life of an animal. This enables representations that factorize across tasks (instead of the task-specific features usually obtained in a continual learning scenario \cite{yang_task_2018}). Afterward, the different tasks can be learned sequentially simply by using task-specific output heads. We call the resulting model a Modular Latent Primitive (MoLaP) network. 

Altogether, we show that by leveraging aplastic functional modules fitted on a corpus' shared statistics, our system is able to sequentially learn multiple tasks more efficiently than in typical continual learning approaches \cite{yang_task_2018,duncker_organizing_2020}. Also, we demonstrate that our model is more robust to perturbations: a desirable property that biological organisms are remarkably good at \cite{basile_preserved_2020,curtis_effects_2004,fu_aging-dependent_2015,westlye_life-span_2010}, compared to artificial systems, which are relatively sensitive \cite{vincent-lamarre_driving_2016}. 
Notably, even including pre-training, our method requires fewer total parameter updates to learn an entire corpus, compared to leading models of biologically plausible multi-task learning.
Finally, we show that generalization to new tasks that share basic structure with the corpus is also facilitated. This draws a path towards cheaper multi-task solutions, and offers new hypotheses for how continual learning might be implemented in the biological brain.

% Implemented in a modular network with streamlined inter-connectivity and using task-specific readouts, this strategy could help avoid task interference and forgetting, while substantially lowering the costs of training.
% Throughout, we argue that this strategy is consistent with experience in a biological organism, complementing structural biases encoded genetically \cite{zador_critique_2019}. 

\section{Relation to prior work}\label{background}

\subsection{Recurrent neural networks}
Similar to the brain, computations in a RNN are implemented by the collective dynamics of a recurrently connected population of neurons \cite{kveraga_top-down_2007}. In fact, when trained on similar tasks as animals, RNNs have been shown to produce dynamics consistent with those observed in animals; they can therefore serve as a proxy for computational studies of the brain \cite{marton_learning_2020,richards_deep_2019,yang_task_2018,remington_dynamical_2018,kell_task-optimized_2018,zeng_continuous_2018,chaisangmongkon_computing_2017,rajan_recurrent_2016,sussillo_neural_2015,mante_context-dependent_2013,sussillo_opening_2013,barak_fixed_2013,buonomano_state-dependent_2009,sussillo_generating_2009}. Thus, by showing the benefits of pre-trained recurrent modules in RNNs, we offer a hypothesis for how multi-task learning is implemented in the brain.

While classic RNNs are trained with backpropagation through time (BPTT), \textit{reservoir computing} approaches such as echo-state networks (ESNs) and liquid-state machines (LSMs) \cite{pathak_model-free_2018,jaeger_harnessing_2004,maass_real-time_2002} avoid \textit{learning} the recurrent weights of the network and instead only learn an output layer on top of a random high-dimensional expansion of the input produced by a random RNN called a \textit{reservoir}. This avoids common problems that come with training RNNs \cite{hochreiter_gradient_2001} and can increase performance while significantly lowering the computational cost \cite{pathak_model-free_2018,jaeger_harnessing_2004,maass_real-time_2002}. %However, reservoir networks struggle to learn complex tasks and they show high susceptibility to local perturbations, even when implemented in a modular way \cite{vincent-lamarre_driving_2016}. 
Our approach mixes both plastic RNNs and reservoir architectures, as modules can be seen as non-random reservoirs initialized to be maximally informative and performing independent computations in isolation.
%However, in the face of partial observability, RNNs trained with BPTT perform better ~\cite{vlachas_backpropagation_2020}

The use of connected modules in RNNs is motivated by the fact that dynamical processes in the world can usually be described causally by the interaction between a few independent and reusable mechanisms \cite{pearl_causality_2009}; modularity in RNNs --- which doesn't emerge naturally \cite{csordas_modular_2020} --- has been shown to increase generalization capabilities \cite{goyal_rims_2019,graves_neural_2014,santoro_rmc_2018, li_modules_2018,henaff_modules_2016,jacobs_experts_1991,jacobs_experts_1991,rosenbaum_modules_2019}. 
%Reservoir computing approaches, such as those proposed in the context of echo-state networks (ESNs) and liquid state machines (LSMs) \cite{pathak_model-free_2018,jaeger_harnessing_2004,maass_real-time_2002}, avoid \textit{learning} the recurrent weights of a network (only training output weights) and thereby achieve reduced training time and lower computational cost compared to recurrent neural networks (RNNs) whose weights are all trained in an end-to-end fashion. This can be advantageous: when full state dynamics are available for training, reservoir networks exhibit higher predictive performance and lower generalisation error at significantly lower training cost, compared to recurrent networks fully trained with gradient descent using backpropagation through time (BPTT); in the face of partial observability, however, RNNs trained with BPTT perform better ~\cite{vlachas_backpropagation_2020}. Also, naive implementations of reservoir networks show high susceptibility to local perturbations, even when implemented in a modular way \cite{vincent-lamarre_driving_2016}.

\subsection{Learning multiple tasks}
In \textit{Multi-Task Learning} (MTL) \cite{crawshaw_mtl_2020}, the model has access to the data of all tasks at all times and must learn them all with the implicit goal of sharing knowledge between tasks. Typically, the model is separated into shared and task-specific parts which are learned end-to-end with gradient descent on all the tasks jointly (batches mix all the tasks).
While we do require information about all tasks simultaneously for our pre-training step, \textit{how} we do so sets us apart from other MTL methods. Indeed, the modules are trained on (the first few canonical components of) the shared computation \textit{statistics} of the tasks rather than directly on tasks themselves; and we argue this offer a more realistic account of optimization pressure animals are subject to over long timescales (e.g., evolution or development). Most previous MTL studies using RNNs focus on NLP or vision tasks. The most similar to our approach is \cite{yang_task_2018}, which uses the same set of cognitive tasks we do, but does not use modules and trains shared parameters end-to-end jointly on all tasks. 

% Maybe not necessary, could cut for space
%  In \textit{Transfer} [cite survey], \textit{Few-shot}[cite survey] and \textit{One-shot}[cite survey] learning, the focus is on leveraging previously trained networks to achieve a reduction in training time on new tasks. Previous works in this direction using RNNs have largely been carried out in time-series forecasting [cite,\cite{orozco_zero-shot_2020}] and in NLP [cite]. Our framework can be interpreted as doing transfer learning on top of the pre-trained module.

%This work, largely carried out in the computer vision domain, has mostly focused on feedforward systems . Recently, a new framework for achieving good time-series forecasts with limited data has been proposed , but it has not been done in a multi-task setting. Initial training in this framework is expensive also, as it requires training on a multitude of different time-series datasets. 

Meanwhile, in \textit{Continual Learning} (CL)  \cite{hadsell_embracing_2020,yu_gradient_2020,adel_continual_2020,parisi_continual_2019,golkar_continual_2019,yoon_lifelong_2018,serra_overcoming_2018,schwarz_progress_2018,zeng_continuous_2018,kirkpatrick_overcoming_2017,zenke_continual_2017,lopez-paz_gradient_2017,draelos_neurogenesis_2017,rusu_progressive_2016,cossu_continual_2020}, the system has to learn many different tasks sequentially and has only access to task-relevant data, one at a time. The main focus of this line of work is to alleviate the hard problem of catastrophic forgetting that comes with training shared representation sequentially \cite{lee_maslow_2022,french_forgetting_1999,ramasesh_anatomy_2020}. We avoid this issue completely by assuming the pre-existence of aplastic modular primitives provided by statistical pre-training; all our shared parameters are frozen. Indeed, once the modules are set, our work can be conceptualized as continual learning since the tasks' output heads are learned sequentially. Note that other CL methods freeze previously learned parameters to avoid interference \cite{rusu_progressive_2016,mallya_packnet_2017}, but they do so one task at a time. Therefore, frozen representations end up being task-specific rather than factorized across tasks like ours; which hurts their efficiency. Closest to our approach is \cite{duncker_organizing_2020} in that they consider the same tasks, however, they learn the shared parameters progressively in orthogonal subspaces to protect them from interference. We use them as our main point of comparison to show the benefits of our approach over typical CL.
%are not readily extensible to dynamical systems where recurrent interactions may lead to unexpected behavior in the evolution of neural responses over time \cite{sodhani_toward_2020}. 

Finally, in \textit{Meta-Learning} \cite{hospedales_meta_2020}, a single model learns to solve a family of tasks by learning on two time-scales. In particular, in gradient-based meta-learning \cite{finn_maml_2017,nichol_reptile_2018}, a network initialization which is good for all tasks is learned over many meta-training steps, speeding up the training of new tasks. The pre-training of the modules on the joint statistics of the corpus in our system has a similar motivation: acquiring primitive latent representations useful across tasks to speed up learning of individual tasks. 

\subsection{Relationship with hierarchical reinforcement learning}
Works in Hierarchical Reinforcement Learning (HRL) \cite{pateria_hrl_2021} also use so-called \textit{primitive} policies in order to solve (sometimes multiple \cite{andreas_hrl_2016}) decision making tasks. In this case, the primitives are complete policies pre-trained to perform some basic action (e.g., move to the right) and they are composed sequentially (in time) to perform complex tasks. In our case, the primitives are combined in a \textit{latent} space by the output layer, thus achieving a different type of compositionality. We also use a supervised learning objective to train our model.

\begin{figure}[hbt!]
    \begin{center}
    \centering
    \hspace*{0cm}\includegraphics[width=1.0\linewidth]{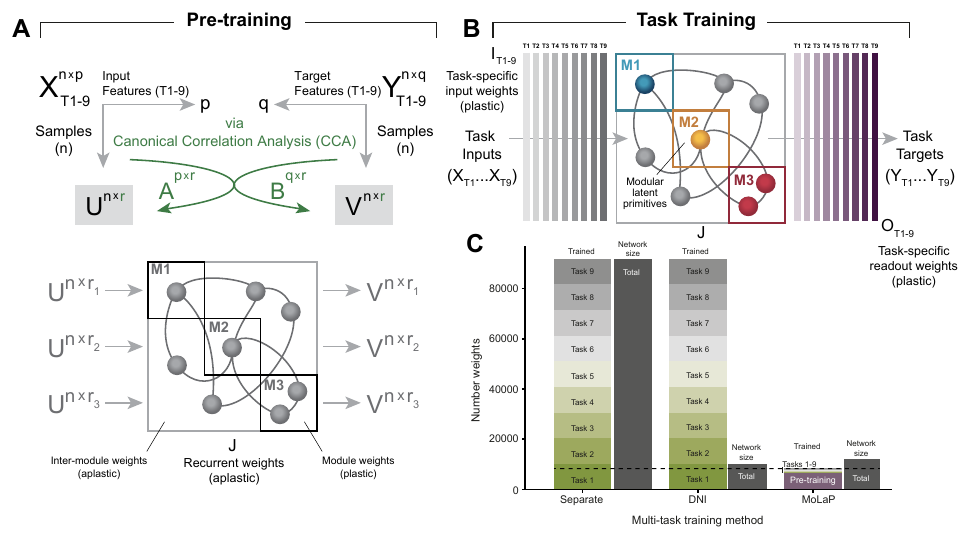}
    \hspace*{0cm}\captionof{figure}{\textbf{Modular latent primitives (MoLaP) network overview}. \textbf{A.} Pre-training leverages canonical correlation analysis (CCA) to derive canonical variables $U$ and $V$ from a corpus of tasks, which are used to pre-train separate network modules. \textbf{B.} During task training, recurrent weights remain frozen and only input and readout weights are adjusted. Task-specific responses can be obtained via separate readout weights. \textbf{C.} Parameter overview for the various multi-task training methods. Parameters trained (or updated) and total network size is compared across the three methods. MoLaP makes $\sim 2$ orders of magnitude fewer parameter updates than DNI, at a relatively moderate increase of total parameter count.}
    \label{fig:Figure1}
    \end{center}
%\end{fullwidth}
\end{figure}
%%%%%%%%%%%%%%%%%%%%%%%%%%%%%%%%%%%%%%%%%%%%%%%%
\section{Model and experimental details}\label{Model details}

\subsection{Modularity}\label{modularity}
Our network is based on an RNN architecture with hidden state $h$ that evolves according to
\begin{equation}\label{eq:Eq.1}
    \begin{aligned}
        \textbf{h}^M_{t+1}\,=\,\sigma(\textbf{W}_{rec}\textbf{h}_{t}\,+\,\textbf{W}_{in}\textbf{x}_{t}\,+\,\boldsymbol{\eta}_{t}),\\
        \newline
        \textbf{y}_t\,=\,\textbf{W}_{out}\textbf{h}^M_{t},\\
    \end{aligned}
\end{equation}
with activation function $\sigma=\tanh$, external input $x^{i}_t$, uncorrelated Gaussian noise $\eta^{i}_{t}$, and linear readouts $y^{i}_t$. Inputs enter through shared weights and outputs are read out through dedicated weights.

Modules are embedded within the larger network with module size $N^{M}\,=\,N^{Total}/{n^{M}}$, total number of units in the network $N^{Total}=100$, and number of modules $n^{M}=2$ throughout this paper. Modules are inter-connected with the reservoir, but not directly to each other (\autoref{fig:Figure1}A). Starting from an initial state, with weights drawn from a Gaussian distribution, each module is trained to produce a module-specific target output time series $\textbf{z}_t$ by minimizing the squared loss using the Adam optimizer (see \ref{appendix} for more details). Module-specific inputs and targets are obtained via canonical correlation analysis (CCA; \autoref{fig:Figure1}A; see section \ref{sec:latent_prim}).

In the \textit{pre-training} phase (\autoref{fig:Figure1}A), \textit{intra}-module weights are plastic (unfrozen) while \textit{inter}-module weights remain aplastic (frozen). Modules are trained sequentially on latent primitives (see Section~\ref{sec:latent_prim}) such that only intra-module connections of the module currently being trained are allowed to change, while the others remain frozen. The gradient of the error, however, is computed across the whole network. Thus for the case outlined in \autoref{fig:Figure1}, for example, module 1 (M1) is trained first, followed by modules 2 (M2) and 3 (M3). When M1 is being trained, only connections inside of M1 are allowed to change, while the connections in the rest of the network remain frozen. After that, M1 connections are frozen, and M2's intra-module connections are unfrozen for training.

\subsection{Tasks}
Neuroscience tasks \cite{yang_task_2018}, like the ones we employ here (explained in greater detail in \ref{tasks}), are structured as trials, analogous to separate training batches. In each trial, stimuli are randomly drawn, e.g., from a distribution of possible orientations. In a given trial, there is an initial period without a stimulus in which animals are required to maintain fixation. Then, in the stimulus period, a stimulus appears for a given duration and the animal is again required to maintain fixation without responding. Finally, in the response period, the animal is allowed to make a response.

\subsection{Deriving latent primitives}
\label{sec:latent_prim}
We derive inputs and targets for \textit{pre-training} directly from the entire corpus of tasks by leveraging canonical correlation analysis (CCA; \autoref{fig:Figure1}A):  

\begin{equation}\label{eq:Eq.2}
    \begin{aligned}
        (\textbf{A}^{'},\textbf{B}^{'})\,=\,\argmax_{\textbf{A},\textbf{B}}\,corr(\textbf{A}^{\intercal}\textbf{X}^{T1-9},\textbf{B}^{\intercal}\textbf{Y}^{T1-9})\\
        \newline
        \textbf{U}_{n\,x\,r}\,=\,\textbf{A}^{'\intercal}\textbf{X}\, \hspace{2.4cm} \, \textbf{V}_{n\,x\,r}\,=\,\textbf{B}^{'\intercal}\textbf{Y}\\
    \end{aligned}
\end{equation}

In CCA, we seek $\textbf{A}$ and $\textbf{B}$ such that the correlation between the two variables $\textbf{X}$ (task inputs) and $\textbf{Y}$ (task outputs) is maximized. We then obtain the canonical variates $\textbf{U}$ and $\textbf{V}$ by projecting $\textbf{A}$ and $\textbf{B}$ back into the sample space of inputs and outputs, respectively, with $n$: number samples and $r$: number of latent dimensions. The canonical variates are used to pre-train the modules (\ref{modularity}), with latent dimensions $r=1,2,3,..$ corresponding to module numbers $M=1,2,3,...$ (\autoref{fig:Figure1}A).

% Taking inspiration from this task structure, and core functionality in most neuroscience tasks \cite{maheswaranathan_universality_2019, gallego_cortical_2018, mante_context-dependent_2013}, we derive simple {\it task primitives} that provide core functionality for solving multiple tasks. The goal is to explore how such primitives can be pre-learned and recombined by networks for multi-task learning. Chosen primitives are (i) a low-dimensional latent representation of inputs, and (ii) the memorization of transient input sequences. In our model, M1 is assigned to (i) and learns to autoencode the fixation signal. M2 and M3 are assigned to (ii) and are trained to hold Inputs 1-2 in memory, respectively. More specifically, M2 receives short positive input pulses ($\Delta t=25ms$) and is required to hold the signal in memory for the remainder
% of the trial period, while M3 is required to do the same with negative input pulses. More details can be found in section \autoref{appendix}.

\subsection{Task training}
In the task-training phase (\autoref{fig:Figure1}B), recurrent weights remain aplastic (like in \textit{reservoir computing}) and only input and output weights are allowed to change. 
Task-specific input ($\textbf{W}_{in}$) and readout weights ($\textbf{W}_{out}$) are trained by minimizing a task-specifc loss function (Eq \ref{eq:Eq.1}). Separate linear readouts are used to obtain the outputs of the network (see \ref{Training details} for details):
\begin{equation}
    \begin{aligned}\label{eq:Eq.3}
        \textbf{y}^{T1-9}_t\,=\,\textbf{W}^{T1-9}_{out}\textbf{h}^M_{t}\\
    \end{aligned}
\end{equation}

\section{Results}\label{results}

%In comparison to other approaches, our modular latent primitives (MoLaP) network achieves similar performance with $\sim$two orders of magnitude less parameter updates (including the pre-training phase) and only a small increase in the total parameter count (\autoref{fig:Figure1}C). 
Training multiple tasks with separate networks requires a high total number of parameters (as separate networks are used for each task) as well as a high number of parameter updates (as all of the network parameters are updated for each task). Thus, assuming a corpus of 9 tasks, a network size of 100 units, one input and one readout unit, naively training separate networks takes $9.18e5$ parameters and parameter updates (\autoref{parameters-table}-Separate networks).
In contrast, using recently proposed {\it dynamical non-interference} (DNI;\cite{duncker_organizing_2020}), one network is trained to perform all tasks; however, all of the parameters still need to be updated on each task as the shared representations are learned sequentially. Thus, DNI still requires $9.18e5$ parameters to be optimized (\autoref{parameters-table}-DNI). 

Similar to DNI, our approach (MoLaP) uses one network to perform all tasks; but MoLaP keeps both the total number of parameters, and the number of parameter updates low (\autoref{fig:Figure1}C-MoLaP). With MoLaP, the number of necessary parameter updates is $\sim2$ orders of magnitude smaller compared to DNI and independent network approaches, with only a relatively modest $18\%$ increase in total parameter count compared to DNI (\autoref{fig:Figure1}C, \autoref{parameters-table}). This arises from using shared reservoirs pre-trained on the joint statistics of the task corpus.

% The recurrent weights of the modular network only need to be trained once upfront, and subsequently, each task can be learnt
% training separate readout weights only. This allows the parameter count in our approach to increase much more slowly as a function of tasks trained (\autoref{fig:Figure4}A) compared to other approaches, which require the entire recurrent weight matrix to be re-trained on each task. The overall parameter count increases slowly with tasks, but also stays low overall using our approach (\autoref{fig:Figure4}B).
%%%%%%%%%%%%%%%%%%%%%%%%%%%%%%%%%%%%%%%%%%%%%%%%%%%%%%%%%%%
% \begin{table}[hbt!]
%   \caption{Parameter count}
%   \label{parameters-table}
%   \centering
%   \begin{tabular}{llll}
%     \toprule
%     % \multicolumn{4}{c}{parts}                   \\
%     % \cmidrule(r){1-4}
%     Counts  & MoLaP & DNI & Separate networks\\
%     \midrule
%     Overall count & 1.2e5 & 1.02e5 & 9.18e5\     \\
%     Number trained & \textbf{1.2e5} & 9.18e5 & 9.18e5 \\
%     Number trained per task & \textbf{2e2}  & 1.02e5 & 1.02e5 \\
%     \bottomrule
%   \end{tabular}
% \end{table}

\begin{table}[hbt!]
  \caption{Parameter count}
  \label{parameters-table}
  \centering
  \begin{tabular}{llll}
    \toprule
    % \multicolumn{4}{c}{parts}                   \\
    % \cmidrule(r){1-4}
    Counts  & Separate networks & DNI & MoLaP\\
    \midrule
    Overall count & 9.18e5 & 1.02e5 & 1.2e5\     \\
    Number trained & 9.18e5 & 9.18e5 & \textbf{5.2e3} \\
    Number trained per task & 1.02e5 & 1.02e5 & \textbf{2e2} \\
    \bottomrule
  \end{tabular}
\end{table}
%%%%%%%%%%%%%%%%%%%%%%%%%%%%%%%%%%%%%%%%%%%%%%%%%%%%%%%%%%%

%%%%%%%%%%%%%%%%%%%%%%%%%%%%%%%%%%%%%%%%%%%%%%%%%%%%%%%%%%%%%%%%%%%%%
%Figure 3: Performance
\begin{figure}
    %\begin{center}
    %\centering
    %\hspace*{-2cm}
    \hspace*{.5cm}\includegraphics[width=.90\linewidth]{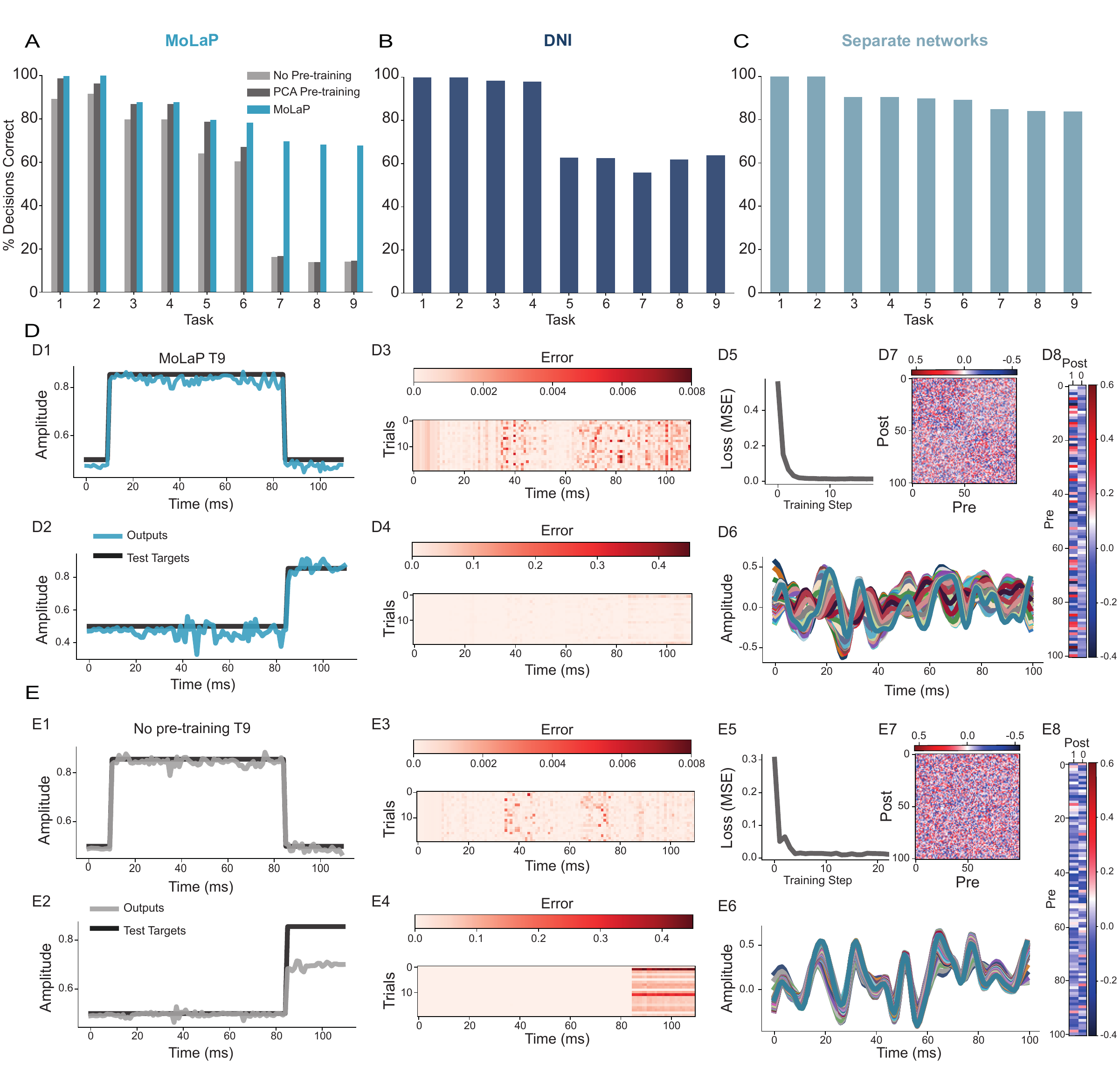}
    \hspace*{.5cm}\captionof{figure}{\textbf{Performance comparison.} Performance ($\%$ decisions correct on a $2000$-trial test set) shown for \textbf{A.} separately trained networks, \textbf{B.} a network trained with  Dynamical non-interference (DNI), and \textbf{C.} with  modular latent primitives (MoLaP). MoLaP performance is also compared to a version without pre-training (where all the connections in the reservoir remain frozen during pre-training) and to pre-training leveraging principal component analysis (PCA). Performance on task T7 for MoLaP with pre-training (\textbf{D.}) and without (\textbf{E}). Outputs vs Targets for output units 1 (\textbf{D1/E1}) and 2 (\textbf{D2/E2}), error (MSE) on 20 examples from the test set for output units 1 (\textbf{D3/E3}) and 2 (\textbf{D4/E4}), training loss (\textbf{D5/E5}), hidden units activity for a sample trial (\textbf{D6/E6}), recurrent weight matrix after training (\textbf{D7/E7}) and output weights after training (\textbf{D8/E8}). See \ref{appendix} for further details.}
    \label{fig:Figure3}
    %\end{center}
%\end{fullwidth}
\end{figure}
%%%%%%%%%%%%%%%%%%%%%%%%%%%%%%%%%%%%%%%%%%%%%%%%%%%%%%%%%%%%%%%%%%%%%%%%%

Despite low parameter update requirements, using MoLaP, we achieve similar performance to both independent networks and DNI (\autoref{fig:Figure3}), with successful learning ($>70\%$ decisions correct) across all 9 tasks (\autoref{fig:Figure3}A,B).
% , comparable to that achieved by another CL approach, dynamical non-interference (DNI; \autoref{fig:Figure3}B; \cite{duncker_organizing_2020}). 
% Training separate networks for each task (\autoref{fig:Figure3}C) achieves slightly better performance ($>80\%$ decisions correct) compared to DNI and MoLaP multi-task learning approaches. 
To further investigate the role of latent primitive representations in MoLaP, we also compared  a MoLaP version without pre-training (using randomly initialized modules like in \textit{reservoir computing}), and a version using principal components analysis (PCA) instead of CCA to obtain latent inputs and targets for pre-training (\autoref{fig:Figure3}A). PCA offers an alternative way to derive latent primitives, and differs from CCA by successively maximizing the variance in the latent dimensions separately for inputs and outputs. The distinction is that PCA extracts a low-dimensional space from a single dataset (the covariance matrix), while CCA explicitly extracts modes that maximize variance \textit{across} variables (the cross-covariance matrix; see also \ref{appendix} for further details). This feature turns out to be crucial for deriving useful latent primitives.

Performance is more similar for easier tasks (T1-T4) than for more complex tasks (T5-T9), where flexible input-output mapping is required \cite{dubreuil_structure_2021}. However, only MoLaP achieves the highest performance across all tasks. We show the performance of the trained MoLaP network \textit{with} pre-training (\autoref{fig:Figure3}D) and \textit{without} (\autoref{fig:Figure3}E) on task T7 (context-dependent memory task 1, see \ref{tasks} for task details). With pre-training, both readout variables hit the target (\autoref{fig:Figure3}Da-b) and decisions can be decoded correctly across the test set (\autoref{fig:Figure3}Dc-d), while in the version without pre-training, one of the readout variables misses the target (\autoref{fig:Figure3}Eb) and thus leads to incorrect decisions (\autoref{fig:Figure3}Ed).

\subsection{Perturbations}

%%%%%%%%%%%%%%%%%%%%%%%%%%%%%%%%%%%%%%%%%%%%%%%%%%%%%%%%%%%
%Figure 4: Modular weight perturbations
\begin{figure}
    %\begin{center}
    %\centering
    \hspace*{0.5cm}\includegraphics[width=.9\linewidth]{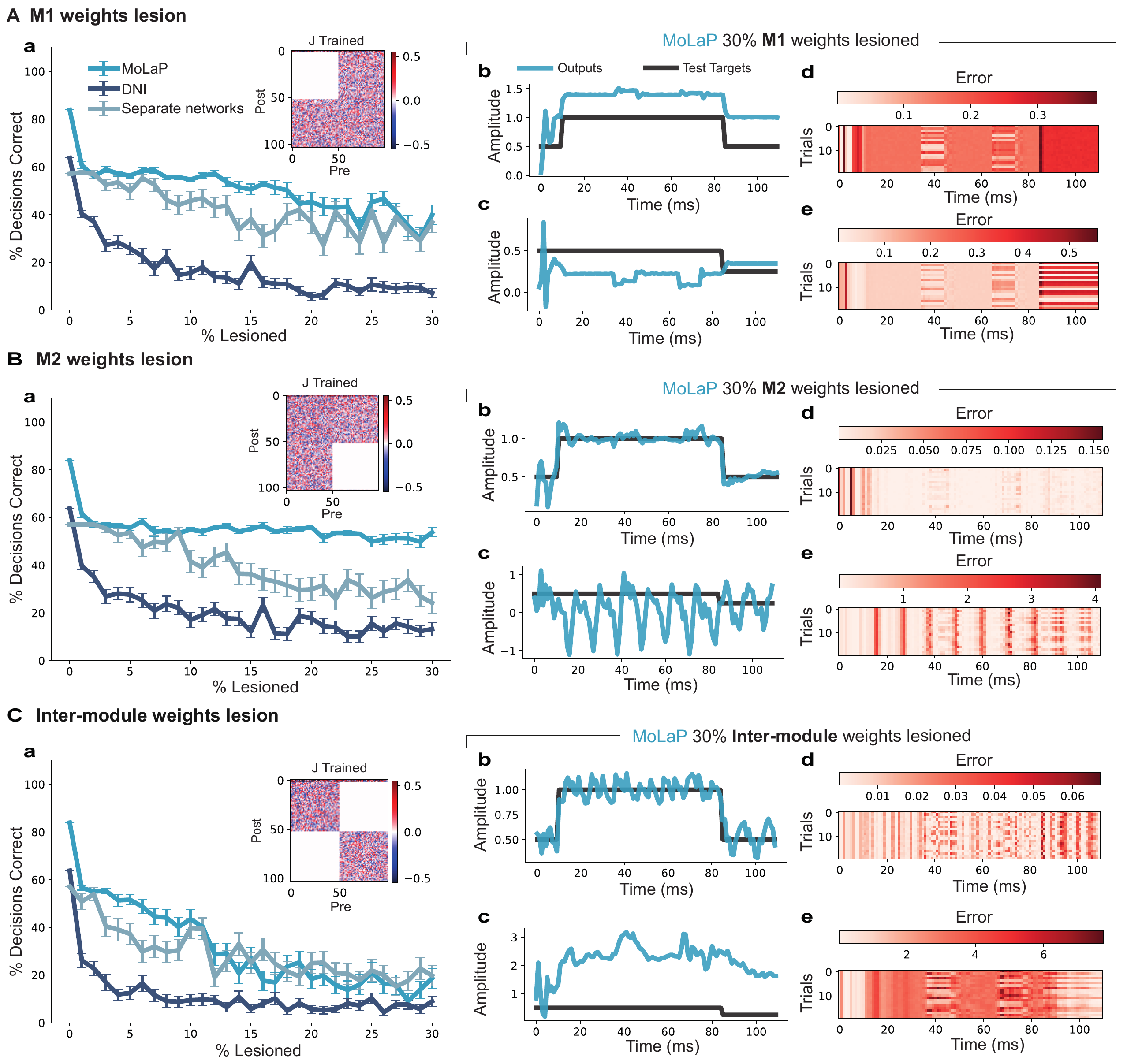}
    \hspace*{0.5cm}\captionof{figure}{\textbf{Performance after modular perturbations.} Performance ($\%$ decisions correct on a 2000 trial test set) after lesioning module 1 (M1) weights (\textbf{A}), module 2 (M2) weights (\textbf{B}), and inter-module weights (\textbf{C}). Performance as a function of percentage lesioned (\textbf{a}), with error bars 
    showing the standard deviation of the mean across 30 random draws. MoLaP performance at $30\%$ weights lesioned (\textbf{b-e}), with outputs vs. targets for output units 1 (\textbf{b}) and 2 (\textbf{c}), and error (MSE) on 20 trials from the test set for output units 1 (\textbf{d}) and 2 (\textbf{e}). See \ref{appendix} for further details.}
    \label{fig:Figure4}
    %\end{center}
%\end{fullwidth}
\end{figure}
%%%%%%%%%%%%%%%%%%%%%%%%%%%%%%%%%%%%%%%%%%%%%%%%%%%%%%%%%%%
%%%%%%%%%%%%%%%%%%%%%%%%%%%%%%%%%%%%%%%%%%%%%%%%%%%%%%%%%%%
%Figure 5: Mean perturbation results across all tasks
\begin{figure}
    %\begin{center}
    %\centering
    \hspace*{0.5cm}\includegraphics[width=0.9\linewidth]{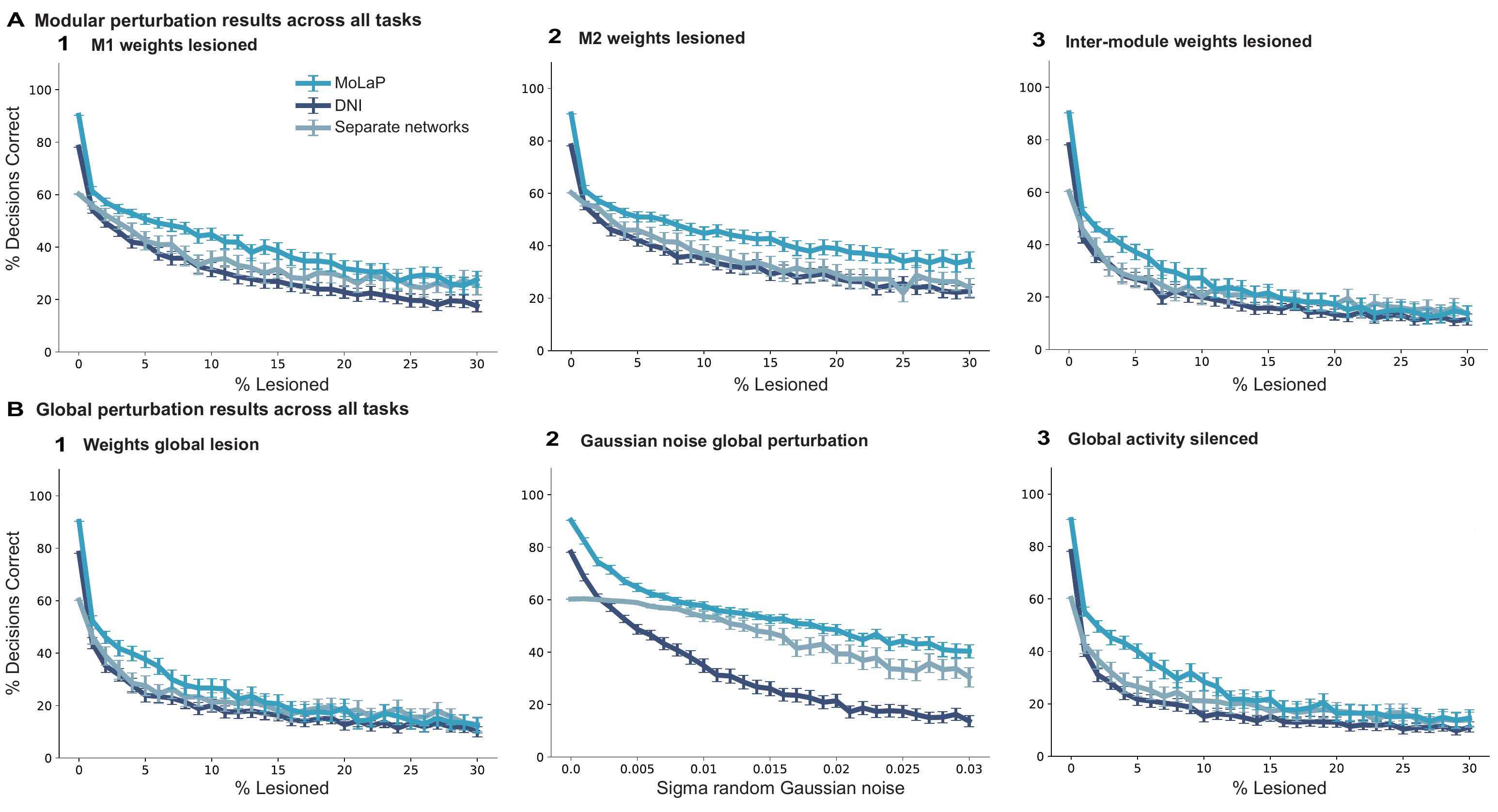}
    \hspace*{0.5cm}\captionof{figure}{\textbf{Performance across all tasks}. Performance ($\%$ decisions correct on a 2000 trial test set) as a function of percentage module 1 (M1) weights lesioned (\textbf{A1}), module 2 (M2) weights lesioned (\textbf{A2}), inter-module weights lesioned (\textbf{A3}), weights lesioned globally (\textbf{B1}), sigma of Gaussian noise added globally (\textbf{B2}, and percentage of units with activity silenced across the entire network (\textbf{B3}). Performance is computed across all nine tasks in the corpus}
    \label{fig:Figure5}
    %\end{center}
%\end{fullwidth}
\end{figure}
%%%%%%%%%%%%%%%%%%%%%%%%%%%%%%%%%%%%%%%%%%%%%%%%%%%%%%%%%%%
We analyzed performance after modular weight lesion for the different approaches, here shown for task T9 (\autoref{fig:Figure4}; see \autoref{fig:SupFigure1}\&\autoref{fig:SupFigure2} for a complete overview of tasks T1-T9)). Weights are ''lesioned" by picking a percentage of entries $W_{rec}^{i,j}$ from the connectivity matrix (\autoref{eq:Eq.1}) and setting them to zero. With module 1 (M1) weights lesioned, MoLaP performance remained high when $<10\%$ of the intra-module weights were lesioned, above that of DNI and comparable to separately trained networks (\autoref{fig:Figure4}A). With module 2 (M2) weights lesioned, MoLaP performance remained high even at a higher percentage of intra-module weights lesioned, above that of DNI and separate networks (\autoref{fig:Figure4}B). When plotting outputs of MoLaP at $30\%$ of the M2 weights lesioned, we observed that the output signal of output unit 1 remained high, resulting in a higher percentage of correctly decoded decisions on average (\autoref{fig:Figure4}Bb,d) compared to when M1 was lesioned (\autoref{fig:Figure4}Ab,d). This can be explained by the fact that the first dimension of the canonical variables (on which M1 was trained) captures a higher percentage of the variance between inputs and targets than the second dimension (on which M2 was trained). Hence, when M2 is lesioned, M1 is still able to compensate and performance is preserved.

We also considered the effect of lesioning \textit{inter}-module connections (\autoref{fig:Figure4}C). Performance degrades quickly when inter-weights are lesioned, for all approaches. When plotting outputs of MoLaP at $30\%$ inter-module connections lesioned, we observed that the combined error across output variables is higher compared to lesioning intra-module weights (\autoref{fig:Figure4}Cb-e). Based on this we would predict that cutting connections between areas is far more devastating than killing a few units within an area.

Furthermore, we analyzed performance after a number of different global perturbations (\autoref{fig:SupFigure3},\autoref{fig:SupFigure4},\autoref{fig:SupFigure5}). We considered three different global perturbations: lesioning entries in the connectivity matrix, adding global Gaussian noise to the entire connectivity matrix, and silencing activity of units in the network (by clamping their activity at zero). All perturbations are repeated for 30 random draws.

We found that on simpler tasks such as T1 (\autoref{fig:SupFigure5}A), all approaches perform similarly for the various global perturbations (see \autoref{fig:SupFigure3}\&\autoref{fig:SupFigure4} for all tasks). DNI performed somewhat better than the other approaches on simpler tasks when weights and activity were lesioned (\autoref{fig:SupFigure5}A,C), and somewhat worse than other approaches when noise was added globally (\autoref{fig:SupFigure5}B). On more complex tasks such as T7 (\autoref{fig:SupFigure5}B), however, we found that MoLaP was more robust to global perturbations. MoLaP performance remained higher compared to other approaches when up to $10\%$ of connections were lesioned (\autoref{fig:SupFigure5}Ba) or when up to $10\%$ of units were silenced (\autoref{fig:Figure5}Bc). MoLaP performance was also robust to global noise, even at relatively high levels (\autoref{fig:SupFigure5}Bb).

%%%%%%%%%%%%%%%%%%%%%%%%%%%%%%%%%%%%%%%%%%%%%%%%%%%%%%%%%%%
%Figure 6: Leave-out performance
\begin{figure}[hbt!]
    \begin{center}
    \centering
    \includegraphics[width=.55\linewidth]{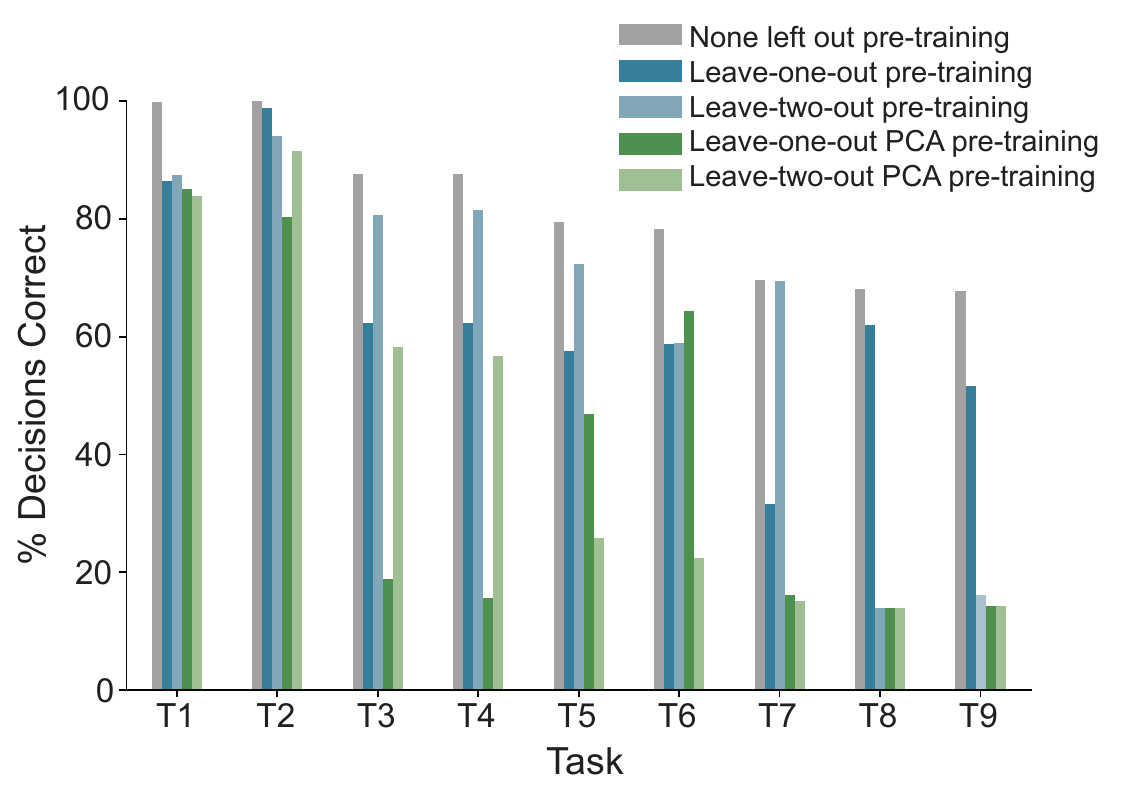}
    \captionof{figure}{\textbf{Generalization performance.} Performance ($\%$ decisions correct on a $2000$-trial test set) for all tasks, after pre-training with CCA where all tasks were included (grey), pre-training with CCA where one task (dark blue) or two (light blue) tasks were left out, and pre-training with PCA where one task (dark green) or two tasks (light green) were left out. For leave-one-out, the left-out task corresponds to the task depicted on the x-axis. For leave-two-out, the task depicted and the one immediately next on the right were left out (in the case of T9, both T9 and T1 were left out).}
    \label{fig:Figure6}
    \end{center}
%\end{fullwidth}
\end{figure}
%%%%%%%%%%%%%%%%%%%%%%%%%%%%%%%%%%%%%%%%%%%%%%%%%%%%%%%%%%%

Over the space of all the tasks we consider, we found that MoLaP performed better than the other approaches for all the perturbations under consideration (\autoref{fig:Figure5}).

\subsection{Generalization}
Finally, we analyzed the generalization performance of MoLaP by holding out each tasks, in turn, from the entire corpus used to derive the canonical variables for pre-training (\autoref{fig:Figure6}; see section \ref{Model details} for details). We found that performance generalized to the left-out task relatively well, achieving higher performance than pre-training on the complete corpus using PCA (see \autoref{fig:Figure3}). We also observed some task-specific effects; performance on T7 was relatively worse when it was left out from the corpus of tasks used in pre-training, compared to when other tasks were left out. Even when two tasks were left out, MoLaP dynamics still generalized well with performance remaining high on the majority of tasks. See Appendix \ref{appendix} for details on 2-tasks hold out protocol.

% Overall, we found that our multi-task learning technique (MoLaP) achieved high robustness to a wide range of perturbations (modular and global). When higher order modules were damaged (beyond M1), but M1 was left intact, we found that performance for some of the output units could be maintained. 

\section{Discussion}\label{Discussion}
We present a new approach for multi-task learning leveraging pre-trained modular latent primitives that act as inductive biases acquired through evolution or previous experience on shared statistics of a task corpus. The modules are inspired from organization of brain activity and provide useful dynamics for 
solving a number of different tasks; including those that are not in the corpus used for pre-training, but share some structure with it (i.e.,~generalization). We showed that our approach achieves high performance across all tasks, while not suffering from the effects of task order or task interference that might
arise with other continual learning approaches \cite{duncker_organizing_2020,zenke_continual_2017}.
Crucially, our approach achieves high performance with $\sim2$ orders of magnitude fewer parameter updates
compared to other approaches (even including the pre-training phase) at the cost of only a moderate increase in the number of total parameters. MoLaP scales better with tasks, thus reducing overall training cost. 

Our results also show that networks composed of pre-trained task modules are more robust to perturbations, even when perturbations are specifically targeted at particular modules. As readouts may draw upon dynamics across different modules, task performance is not drastically affected by the failure of any one particular module. The lower performance of the fully connected modular network may be due to the trained connection weights, which increase the inter-connectivity of different cell clusters and thus the susceptibility of local cell clusters to perturbations in distal parts of the network.

This result generates useful hypotheses for multi-task learning in the brain. It suggests that decision-making related circuits in the brain, such as in the prefrontal cortex, need not re-adjust all weights during learning.
Rather, pre-existing dynamics may be leveraged to solve new tasks. This prediction could be tested by using recurrent networks fitted to population activity recorded from the brain \cite{perich_inferring_2020,pandarinath_inferring_2018} as reservoirs of useful dynamics. We hypothesize that the fitted networks contain dynamics useful to new tasks; if so, it should be possible to train readout weights to perform new tasks. Moreover, targeted lesions may be performed to study the effect of perturbations on network dynamics in the brain; our work suggests networks with functional task modules are less prone to being disrupted by perturbations. Based on our work we would also expect functional modules to encode a set of \textit{basis}-tasks (commonly occurring dynamics) for the purpose of solving future tasks.

In a recent study, \cite{flesch_rich_2021} examined the effects of 'rich' and 'lazy' learning on acquired task representations and compared their findings to network representations in the brain of macaques and humans. 
The lazy learning paradigm is similar to ours in as far as learning is confined to readout weights; but
it lacks our modular approach and does not show performance on multiple tasks. Conversely, our networks are 
not initialized with high variance weights as in the lazy regime. Ultimately, we observe improved robustness using our approach, unlike lazy learning which appears to confer lower robustness compared with rich learning.

This work enhances our understanding of resource-efficient continual learning in recurrent neural networks. As such, we do not anticipate any direct ethical or societal impact. Over the long-term, our work can have impact on related research communities such as neuroscience and deep learning, with societal impact depending on the development of these fields.

\bibliography{MS_MODULAR_NEURIPS.bib}

% References follow the acknowledgments. Use unnumbered first-level heading for
% the references. Any choice of citation style is acceptable as long as you are
% consistent. It is permissible to reduce the font size to \verb+small+ (9 point)
% when listing the references.
% Note that the Reference section does not count towards the page limit.
\medskip

\newpage 

\newpage

\appendix
\section{Appendix} \label{appendix}

\subsection{Tasks}\label{tasks}
We show that we can leverage our pre-configured modular network to learn 
a set of nine different neuroscience tasks, previously used
to study multi-task learning \cite{yang_task_2018,duncker_organizing_2020}.
Trained simultaneously into one and the same network, these tasks produce a 
rich and varied population structure and are mapped into different parts of state space \cite{yang_task_2018}.
Simultaneous training, however, does not guarantee good performance on all tasks; this can be improved 
by using a multi-task training paradigm whereby new tasks are projected into non-interfering 
subspaces\cite{duncker_organizing_2020}. This approach is still dependent on task \textit{order}, however: 
how well a new task is learnt often depends on its placement in the training curriculum.

We employ a set of nine different tasks: \textit{Delay Pro}, \textit{Delay Anti}, 
\textit{Mem Pro}, \textit{Mem Anti}, \textit{Mem Dm 1}, \textit{Mem Dm 2}, 
\textit{Context Mem Dm 1}, \textit{Context Mem Dm 2} and \textit{Multi Mem}.
In \textit{Delay Pro} and \textit{Delay Anti}, the network receives three inputs (Fixation, cos($\theta$), sin($\theta$), 
with $\theta$ varying from 0-180 degrees). The angle inputs turned on at the Go-signal and remained on for
the duration of the trial. The network was required to produce three outputs, yielding the correct location
during the output period (when the fixation signal switched to zero). In the \textit{anti}-versions, the network
is required to respond in the direction opposite to the inputs. In \textit{Mem Pro} and \textit{Mem Anti}, 
the input stimulus turns off after a short, variable duration. As before,
the network is required to produce three outputs encoding the correct location during the output period.

In \textit{Mem Dm 1} and \textit{Mem Dm 2}, the network receives two inputs (Fixation, and a set of stimulus 
pulses separate by 100msec). The network is required to reproduce the higher
valued pulse during the output period. Depending on the task, the network receives one or the other input stimulus pulse set.

In \textit{Context Mem Dm 1} and \textit{Context Mem Dm 2}, the network receives three inputs (Fixation, and two sets 
of stimulus pulses). The network is required to reproduce the higher value pulse during the output period, ignoring the other input stimulus set. The two tasks differ in which input stimulus to pay attention to. 
In \textit{Multi Mem}, the network again receives three inputs (Fixation, and both sets of stimulus pulses). 
The network needs to yield the summed value of the stimulus train with the higher total magnitude of the two pulses.

In models, a context signal indicates when to maintain fixation, separate inputs are fed in through different channels with dedicated input weights, and the choice can be indicated (and decoded) from separate output units for each target.

\subsection{Training details}\label{Training details}
All networks were trained on a single laptop with the Apple M1 Max processor using a custom implementation of recurrent neural networks based on JAX \cite{jax2018github}. The code to train MoLaP networks can be found at \url{https://github.com/dashirn/MoLaP}. We used a recurrent neural network (RNN) model with the hyperbolic tangent as the activation function throughout ($\sigma=\tanh$, \autoref{eq:Eq.1}), in accordance with previous approaches to modeling neuroscience tasks \cite{marton_learning_2020,chaisangmongkon_computing_2017,mante_context-dependent_2013}. Previous work suggests the dynamics obtained with different activation functions, such as $\tanh$ or $ReLu$, and different network models such as RNNs, GRUs and LSTMs are similar \cite{maheswaranathan_universality_2019}. All networks were trained by minimizing the mean squared error using Adam as an optimizer. We used a custom implementation of DNI (\cite{duncker_organizing_2020}) leveraging the same setup as above based on JAX; we also compared our results with those obtained from the native implementation (available at \url{https://github.com/LDlabs/seqMultiTaskRNN} under a MIT license), and obtained similar results.

We performed a gridsearch over various parameter ranges (\autoref{trainparams-table}), and settled on the parameter combination that worked best across all training paradigms. We used a network of $N=100$ for MoLaP and separate networks, and $N=200$ for DNI. 

% This ensured that individual modules trained well, and achieved the highest task performance across all training paradigms. The performance we obtained agrees with previously obtained results \cite{duncker_organizing_2020}. 

\begin{table}[hbt!]
  \caption{Parameter settings}
  \label{trainparams-table}
  \centering
  \begin{tabular}{llll}
    \toprule
    % \multicolumn{4}{c}{parts}                   \\
    % \cmidrule(r){1-4}
    Parameter  & Parameter range considered\\
    \midrule
    Network size & 20\,--\,300\\
    Learning rate & 1e-3\,--\,1e-1\     \\
    Batch size & 20\,--\,400\\
    Scaling factor $g$ & 1.1\,--\,1.8\\
    $L2$-norm weight regularization parameter & 5e-6\,--\,5e-5\\
    Activity regularization parameter & 1e-7\\
    Number iterations before decreasing learning rate & 100\,--\,300\\
    \bottomrule
  \end{tabular}
\end{table}

\subsubsection{Pre-training}
The canonical correlation analysis (CCA) model was fitted to inputs and outputs across the entire corpus of tasks for the entire training set simultaneously, and then divided into separate batches for training. We found that two latent dimensions in the canonical variables ($U$ and $V$, see \eqref{eq:Eq.2}) captured $\approx 60\%$ of the variance in the relationship between inputs and targets, while three captured $\approx 70\%$. We also fitted MoLaP networks with three modules and found they portrayed similarly to the results shown here with two modules.\\

For pre-training the recurrent network in MoLaP, we set the learning rate at $\alpha=3e-3$, with a batch size of 200 trials, the scaling factor at $g=1.5$, the $L2$-norm weight regularization parameter at 1e-5, the activity regularization parameter at 1e-7, and the number of iterations at 200. Each module was trained separately, with the number of units set at $N^{M1}=N^{M2}=50$. When a particular module was being trained, all other connections in the network remained frozen (information from the error gradient was prohibited from updating frozen units). So, for example, during training of module 1 (M1), all connections except those within module remained frozen. After M1 pre-training was complete, M1 connections were frozen M2 connections were unfrozen.\\

\subsubsection{Pre-training with PCA}
Alternatively, we also used principal components analysis (PCA) to derive inputs and outputs for pre-training. PCA was performed on the entire corpus of tasks for the entire training set simultaneously, separately for inputs and targets across all tasks. We thus obtained input and output vectors in the reduced space which we used analogously to CCA to train each module sequentially. 

\subsubsection{Task-training}
In MoLaP, after pre-training, all connections in the recurrent weight matrix were frozen and only input and readout weights were allowed to change. Task-specific input and readout weights were trained for tasks T1-9.\\

For task-training in MoLaP, we set the learning rate at $\alpha=1e-2$ for all networks, the batch size at 200 trials, the scaling factor at $g=1.5$, the $L2$-norm weight regularization parameter at 1e-5, the activity regularization parameter at 1e-7, and the number of iterations at 200. Task-training in separate networks and with DNI only differed in that we set the learning rate at $\alpha=3e-3$ which we found worked best.\\ 

Across all approaches, the weights in the recurrent weight matrix, $\textbf{W}_{rec}$ (\autoref{eq:Eq.1}), were initially drawn from the standard normal distribution and scaled by a factor of $\frac{g}{\sqrt{N}}$. The input weights, $\textbf{W}_{in}$, were also initially drawn from the standard normal distribution and scaled by a factor of $\frac{1}{\sqrt{N_{in}}}$ with $N_{in}$ varying by task. Each input unit also received an independent white noise input, $\boldsymbol{\eta}^{i}$, with zero mean and a standard deviation of $\sigma=0.002$. Output weights, $\textbf{W}_{out}$, were also initially drawn from the standard normal distribution and multiplied by a factor of $\frac{1}{\sqrt{N_{out}}}$ with $N_{out}$ varying by task. 

\subsection{Task performance}
In order to measure task performance, a test set of 2000 trials was randomly generated from a separate random seed in JAX for each task. Performance was calculated as the percentage of correct decisions over this test set. \textit{Decisions}, rather than other error metrics such as mean squared error, were chosen as a performance metric as they reflect the ultimate goal of biological agents. For all output units other than unit 1, a decision was considered correct if the mean output over the response period did not deviate more than a fixed amount ($\delta=0.1$) from the target direction. For output unit 1 (which indicated if fixation was maintained throughout the task), a deviation of ($\delta=0.1$) was permitted. For all output units, the response was decoded by taking the mean across the last 30 time points (response period) in a given trial. Performance was averaged across all output units.

\subsection{Perturbation studies}
We considered 3 types of perturbations: lesioned weights, silenced activity, and added Gaussian noise. In the case of lesioned weights, perturbations were performed both in a \textit{modular} and in \textit{global} fashion. In modular perturbations, connections to lesion were picked randomly from within modules only, or from among inter-modular connections (\autoref{fig:SupFigure1},\autoref{fig:SupFigure2}). Global perturbations involved all the units in the network (\autoref{fig:SupFigure3}, \autoref{fig:SupFigure4}, \autoref{fig:SupFigure5}). Weight lesioning was performed by setting chosen connections to zero. Activity silencing was performed by silencing the activity of chosen units. Gaussian noise of different variance $\sigma \in [0,0.03]$ was added to the entire connectivity matrix. All perturbations were repeated for 30 different random draws.\\

\subsection{Supplementary results}

\setcounter{figure}{0}
\makeatletter 
\renewcommand{\thefigure}{S\@arabic\c@figure}
\makeatother

\newpage

%%%%%%%%%%%%%%%%%%%%%%%%%%%%%%%%%%%%%%%%%%%%%%%%%%%%
%SupFig1: Modular Perturbations T1-T4
\begin{figure}
    % \begin{center}
    % \centering
    \hspace*{0cm}\includegraphics[width=1.0\linewidth]{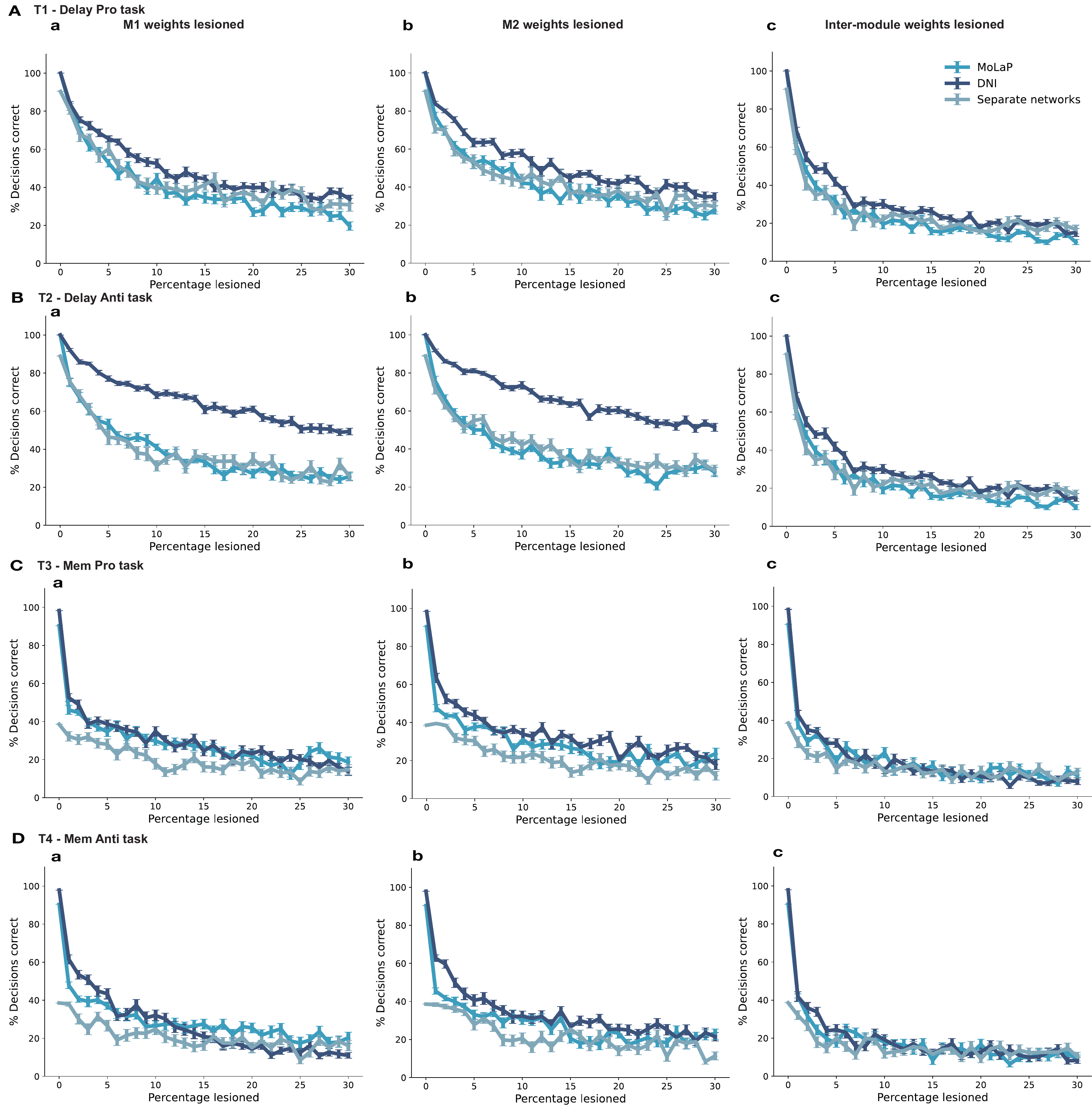}
    \hspace*{0cm}\captionof{figure}{\textbf{Performance after modular weight lesions (T1-T4).} Performance (\% decisions correct on a 250 trial test set) for tasks T1-T4}
    \label{fig:SupFigure1}
    % \end{center}
%\end{fullwidth}
\end{figure}
%%%%%%%%%%%%%%%%%%%%%%%%%%%%%%%%%%%%%%%%%%%%%%%%%%%%
%SupFig2: Modular Perturbations T5-T9
\begin{figure}
    % \begin{center}
    % \centering
    \hspace*{0cm}\includegraphics[width=1\linewidth]{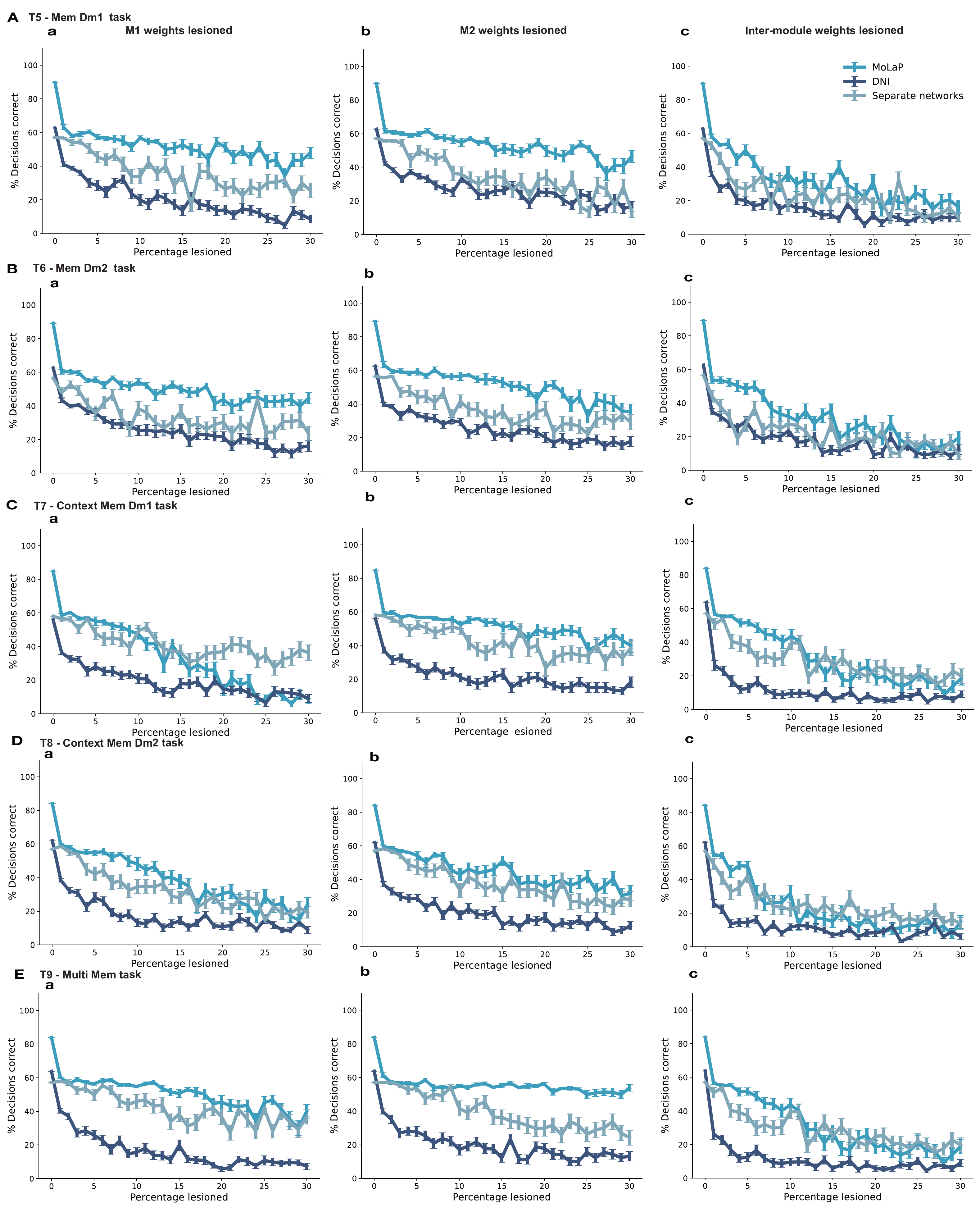}
    \hspace*{0cm}\captionof{figure}{\textbf{Performance after modular weight lesions (T5-T9).} Performance (\% decisions correct on a 250 trial test set) for tasks T1-T4 (\textit{MemDm1}, \textit{MemDm2}, \textit{ContextMemDm1}, \textit{ContextMemDm2}, \textit{MultiMem})}
    \label{fig:SupFigure2}
    % \end{center}
%\end{fullwidth}
\end{figure}
%%%%%%%%%%%%%%%%%%%%%%%%%%%%%%%%%%%%%%%%%%%%%%%%%%%%
%SupFig3: Global Perturbations T1-T4
\begin{figure}
    % \begin{center}
    % \centering
    \hspace*{0cm}\includegraphics[width=1\linewidth]{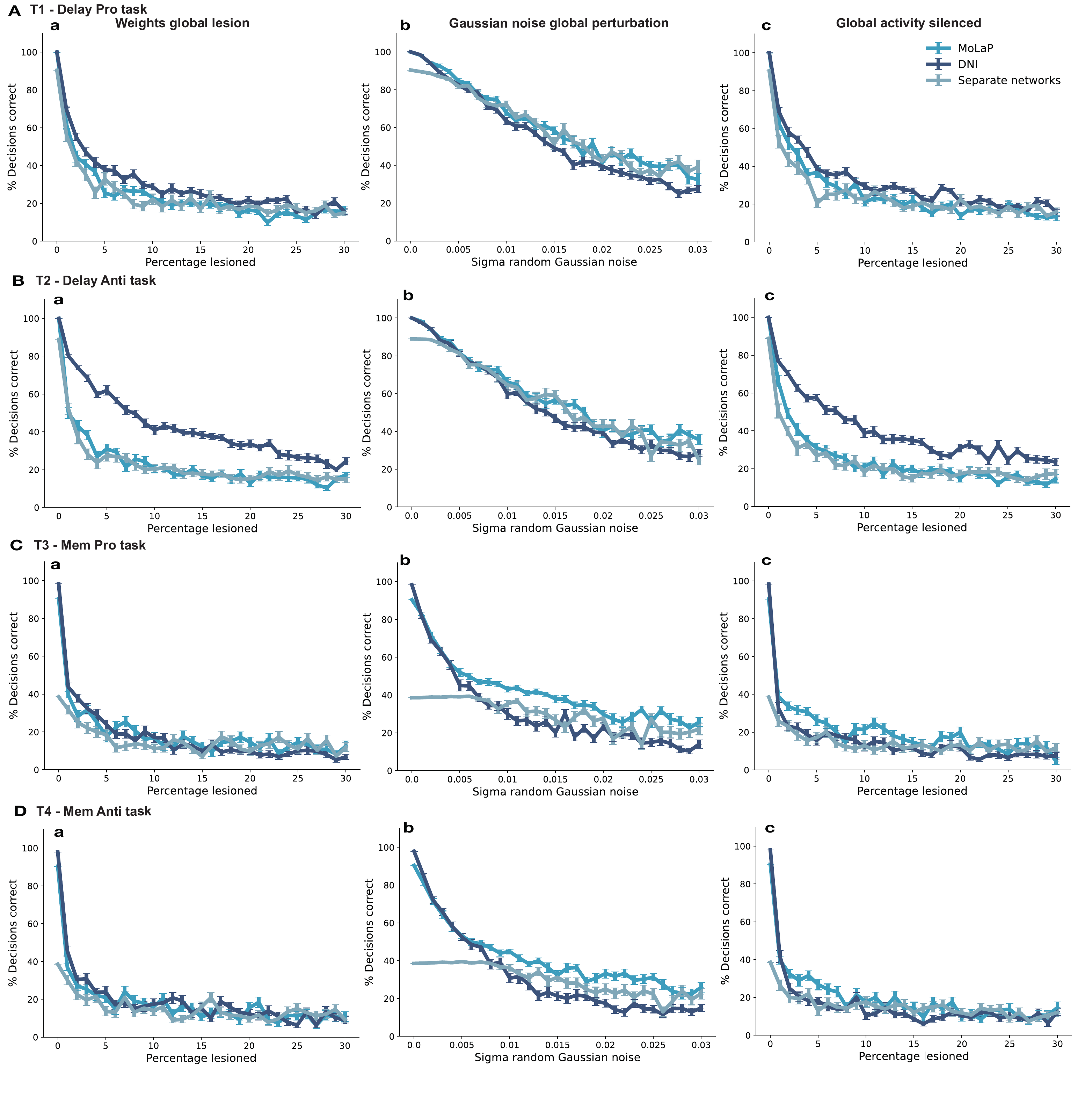}
    \hspace*{0cm}\captionof{figure}{\textbf{Performance after global perturbations (T1-T4).} Performance (\% decisions correct on a 250 trial test set) for tasks T1-T4 (\textit{MemDm1}, \textit{MemDm2}, \textit{ContextMemDm1}, \textit{ContextMemDm2}, \textit{MultiMem}) as a function of \% weights lesioned (A, E, I, M, Q), \% units silenced (B, F, J, N, R),}
    \label{fig:SupFigure3}
    % \end{center}
%\end{fullwidth}
\end{figure}
%%%%%%%%%%%%%%%%%%%%%%%%%%%%%%%%%%%%%%%%%%%%%%%%%%%%
%SupFig4: Global Perturbations T5-T9
\begin{figure}
    % \begin{center}
    % \centering
    \hspace*{0cm}\includegraphics[width=1\linewidth]{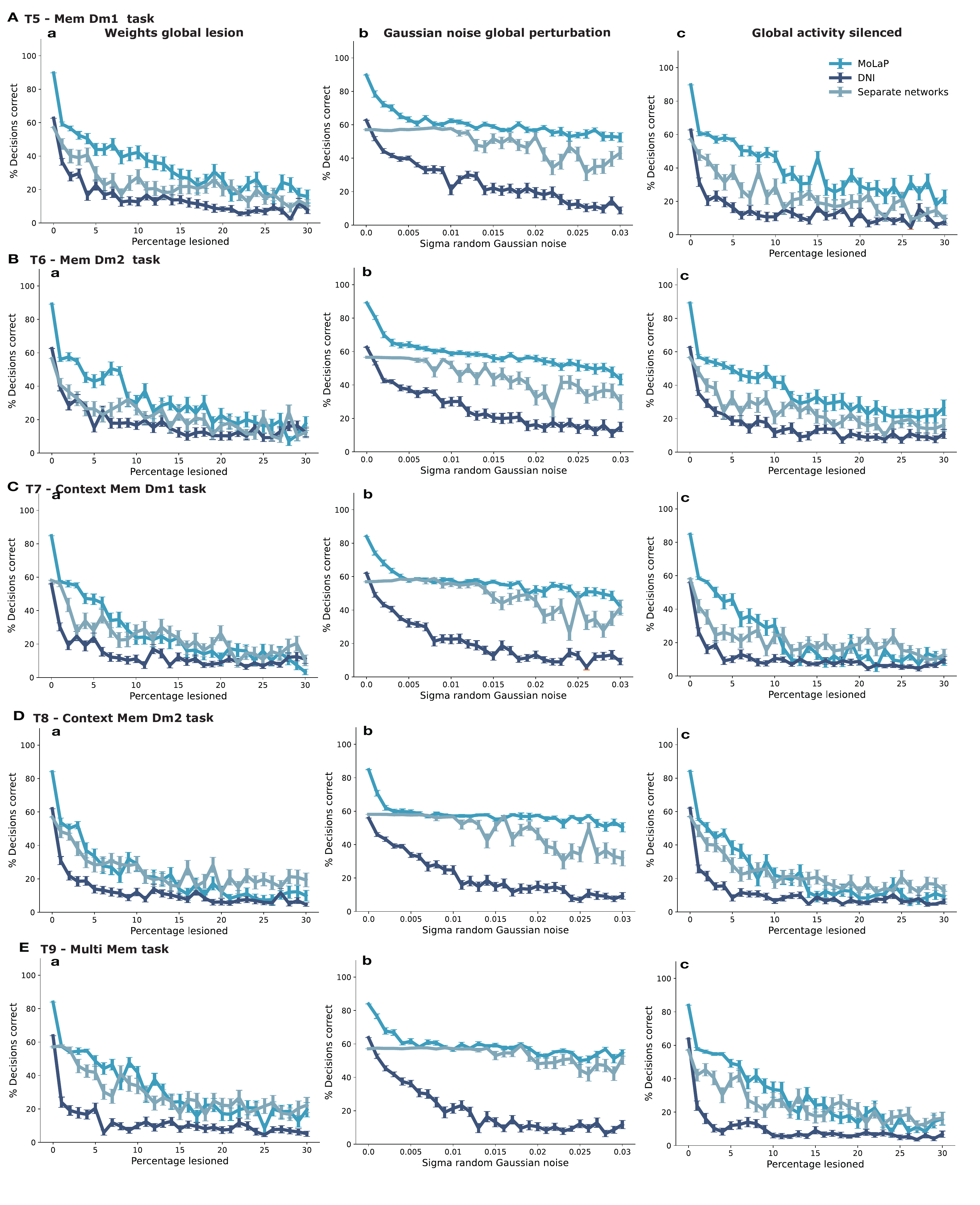}
    \hspace*{0cm}\captionof{figure}{\textbf{Performance after global perturbations (T5-T9)} Performance (\% decisions correct on a 250 trial test set) for tasks T1-T4 (\textit{MemDm1}, \textit{MemDm2}, \textit{ContextMemDm1}, \textit{ContextMemDm2}, \textit{MultiMem})}
    \label{fig:SupFigure4}
    % \end{center}
%\end{fullwidth}
\end{figure}
% Optionally include extra information (complete proofs, additional experiments and plots) in the appendix.
% This section will often be part of the supplemental material.
%%%%%%%%%%%%%%%%%%%%%%%%%%%%%%%%%%%%%%%%%%%%%%%%%%%
%SupFig5: Global Perturbations overview
\begin{figure}
    %\begin{center}
    %\centering
    \hspace*{0.5cm}\includegraphics[width=.9\linewidth]{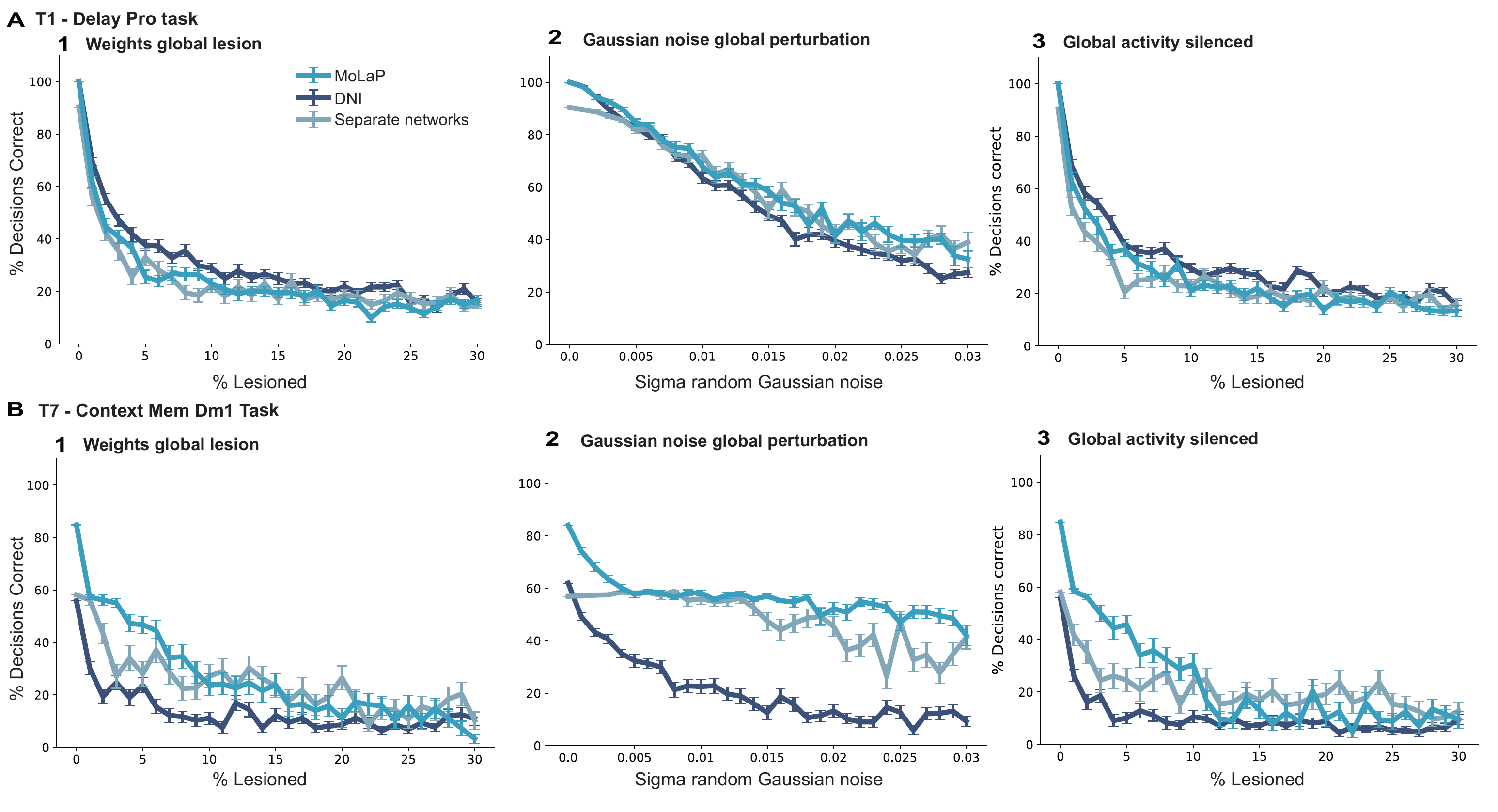}
    \hspace*{0.5cm}\captionof{figure}{\textbf{Performance after global perturbations}. Performance ($\%$ decisions correct on a 2000 trial test set) as a function of percentage weights lesioned globally (\textbf{a}), sigma of Gaussian noise added globally (\textbf{b}, and percentage of units with activity silenced (\textbf{c}), for task T1 (\textbf{A}) and task T7 (\textbf{B}). See \ref{appendix} for further details.}
    \label{fig:SupFigure5}
    %\end{center}
%\end{fullwidth}
\end{figure}

\end{document}